\documentclass[twoside,11pt]{article}

\usepackage{jmlr2e}
\usepackage{graphicx}
\usepackage{booktabs}
\usepackage{subfigure}
\usepackage{amsfonts}
\usepackage{amsmath}
\usepackage{algorithm}
\usepackage{algorithmic}
\usepackage{caption}
\usepackage{tabularx}
\usepackage{amsmath}
\usepackage{epstopdf}
\usepackage{verbatim}
\usepackage{color}
\usepackage{multirow}
\usepackage{url}
\usepackage[table]{xcolor}
\usepackage{multirow}
\newcommand{\calP}{{\cal P}} % changed to q, which is more typical for approximate distributions

\newcommand{\s}[1]{^{(#1)}}
\newfont{\bboard}{msbm10 scaled\magstephalf}

\newcommand{\Normalv}[3]{{\cal N}\left(#1\middle|#2,#3\right)}

\def\EV{\mbox{\bboard E}}

\def\real{\mbox{\bboard R}}

\newcommand{\refeqn}[1]{(\ref{#1})}

\newcommand{\junk}[1]{}

\newcommand{\mbf}[1]{\mathbf{#1}}
\newcommand{\bz}{\mbf{z}}

\newcommand{\bx}{\mbf{x}}

\newcommand{\ba}{\mbf{a}}
\newcommand{\bA}{\mbf{A}}

\newcommand{\bzero}{\mbf{0}}
\newcommand{\bone}{\mbf{1}}
\newcommand{\br}{\mbf{r}}

\newcommand{\bpi}{\boldsymbol{\pi}}

\newcommand{\bmu}{\boldsymbol{\mu}}

\newcommand{\bSigma}{\boldsymbol{\Sigma}}

\newcommand{\detbar}[1]{\left|#1\right|}

\newcommand{\subsubsubsection}[1]{\vspace{0.2cm}\noindent{\bf #1:} }

\newcommand{\CUT}[1]{}

\ShortHeadings{EMHMM Simulation Study}{}
\firstpageno{1}

\begin{document}

\title{EMHMM Simulation Study}

\author{\name Antoni B. Chan \email abchan@cityu.edu.hk \\
       \addr Department of Computer Science\\
       City University of Hong Kong\\
      Kowloon Tong, Hong Kong      
      \\\\
      \name Janet H. Hsiao \email jhsiao@hku.hk \\
       \addr Department of Psychology\\
       The University of Hong Kong\\
       Hong Kong
       }
\editor{}

\maketitle

\begin{abstract}Eye Movement analysis with Hidden Markov Models (EMHMM) is a method for modeling eye fixation sequences using hidden Markov models
(HMMs).  In this report, we run a simulation study to investigate the estimation error for learning HMMs with variational Bayesian inference, with respect to the number of sequences and the sequence lengths. We also relate the estimation error measured by KL divergence and L1-norm to a corresponding distortion in the ground-truth HMM parameters.  From the results of the simulation study, we make recommendations about how many fixation samples are needed to estimate an HMM that is representative of the overall eye gaze strategy of the subject.
\end{abstract}

\vspace{0.5cm}
\noindent
{\small v2: 2017-June-17}

\section{Introduction}

Eye Movement analysis with Hidden Markov Models (EMHMM) \citep{2014jov,emhmm} is a method for modeling eye fixation sequences using hidden Markov models (HMMs).  Given a subject's eye fixation sequences, a subject's HMM is learned using a variational Bayesian approach.  If the HMM is well estimated, then the subject's HMM can then be interpreted as the overall eye gaze strategy of the person.   In this paper we run a simulation study to investigate the estimation error for learning HMMs with respect to the number of sequences and the sequence lengths.  We then make recommendations about how many fixations are required to make the interpretation that the subject's HMM is representative of their underlying eye gaze strategy.
Note that if fewer fixations are used, then it is still okay to estimate HMMs from the data. In this case, the estimated HMM represents the subject's eye gaze pattern on the particular stimuli, not the subject's underlying strategy.  Such HMMs are still valid for further analysis, such as for classification or regression.

The remainder of this paper is organized as follows. In Section \ref{text:background}, we review the HMM and its probability distribution.  In Section \ref{text:simsetup}, we introduce the simulation procedure, including how the HMMs are estimated and how HMMs are compared.  In Sections \ref{text:setup} and \ref{text:results}, we present the experiment setup and the results.  Finally, in Section \ref{text:summary}, we give recommendations on the sample size needed, and conclude.

\section{Background}
\label{text:background}

We consider an HMM with Gaussian emissions. The joint likelihood of the observation (fixations) and the hidden state (ROI) sequences is,
	\begin{align}
	p(\bx, \bz) = p(\bx|\bz)p(\bz) = p(z_1) \prod_{t=2}^{\tau} p(z_{t}|z_{t-1}) \prod_{t=1}^{\tau} p(\bx_{t}|z_{t})
	\label{eqn:HMMjoint}
	\end{align}
where $\bx= [\bx_1, \cdots, \bx_\tau]$ is a sequence of observations and $\bz=[z_1,\cdots z_\tau]$ is the sequence of hidden states, where $\bx_t \in \real^D$ and $z_t \in \{1,\cdots,K\}$. $D$ is the dimension of the observations, and $K$ is the number of states.
Note that for this simulation study, we assume that the length of the sequences is the same for all observations. 

The observation likelihood, transition probabilities, and initial state probabilities are as follows:
	\begin{align}
	\mathrm{initial\ state:}& &
	p(z_{1}=k|\bpi) &= \pi_k
	\label{eqn:HMMinitial}
	\\
	\mathrm{transition\ probability:}& & 
	p(z_{t}=k|z_{t-1}=j) &= a_{j,k}
	\label{eqn:HMMtrans}
	\\
	\mathrm{observation\ likelihood:} & & 
	p(\bx_{t}|z_{t}=k) &=  \Normalv{\bx_{t}}{\bmu_k}{\bSigma_k}
	\label{eqn:HMMlik}
	\end{align}
where the Gaussian density is
	\begin{align}
	 \Normalv{\bx}{\bmu}{\bSigma}
	 = (2\pi)^{-D/2} \detbar{\bSigma}^{-1/2} e^{-\frac{1}{2}(\bx-\bmu)^T \bSigma^{-1}(\bx-\bmu)},
	\end{align}
where $\bSigma$ is the covariance matrix.
The parameters of the HMM are: 
	\begin{align}
	\text{the initial hidden state probabilities:}& & &\bpi = [\pi_1,\cdots,\pi_K]^T  \\
	\text{transition probability $p(z_{t}=k|z_{t-1}=j)$:}& &  &a_{j,k} \\
	\text{transition distribution given $z_{t-1}=j$:}& & &\ba_{j} = [a_{j,1},\cdots,a_{j,k}] \\
	\text{the transition matrix:}& & &\bA=[a_{j,k}]_{j,k} \\
	\text{the mean and covariance matrix for the $k$th Gaussian emission:}& &  
		&\{\bmu_k, \bSigma_k\} 
	\end{align}

The likelihood of an observation sequence $\bx$ is obtained by marginalizing out hidden states, 
	\begin{align}
	p(\bx) = \sum_{\bz} p(\bx|\bz)p(\bz) = \sum_{z_1} \cdots \sum_{z_\tau} p(\bx|\bz)p(\bz)
	\label{eqn:HMMobs}
	\end{align}
The distribution of the initial observation is a Gaussian mixture model (GMM),
	\begin{align}
	p(\bx_1) = \sum_{j=1}^K p(\bx_1|z_1=j)p(z_1=j) =  \sum_{j=1}^K \pi_j p(\bx_1|z_1=j).
	\end{align}
	
\CUT{
As time progresses, the HMM enters a {\em steady-state} distribution, where the state probability stabilizes,
	\begin{align}
	p^*(z_t) = \lim_{t \rightarrow \infty} (\bA^T)^t \bpi \triangleq \bpi^*.
	\end{align}
The steady-state observation distribution is then a GMM, 
	\begin{align}
	p^*(\bx_t) = \sum_{j=1}^K p(\bx_t|z_t=j)p^*(z_t=j) = \sum_{j=1}^K \pi^*_j \Normalv{\bx_t}{\bmu_j}{\bSigma_j}.
	\end{align}
Finally, we can compute a steady-state joint observation distribution over a pair of observations $\bx_{t-1}$ and $\bx_{t}$, 
	\begin{align}
	p^*(\bx_{t-1}, \bx_t) &= \sum_{k=1}^K \sum_{j=1}^K p(\bx_{t-1}|z_{t-1}=j) p(\bx_{t}|z_t=k) p^*(z_{t-1}=j, z_t=k) \\
	&= \sum_{j,k} (a_{j,k} \pi^*_j)  \Normalv{\begin{bmatrix}\bx_{t-1}\\\bx_{t}\end{bmatrix}}{\begin{bmatrix}\bmu_j\\ \bmu_k\end{bmatrix}}{\begin{bmatrix}\bSigma_j & \bzero \\ \bzero & \bSigma_k \end{bmatrix}}, 
	\end{align}
which is a GMM with $K^2$ components.
}

\section{Simulation Procedure}
\label{text:simsetup}

We next outline the simulation procedure for EMHMM. Given a ground-truth HMM, we sample a number of fixation sequences of a given length.  Using the sample sequences, we then estimate an HMM using EMHMM, and then compare the estimated HMM with the ground-truth HMM in terms of various metrics.

\subsection{Ground-truth HMMs}

A ground-truth HMMs $\Theta$ is created from the fixation sequences collected from a subjects.  An HMM was learned from the subject's data using the EMHMM toolbox, and this HMM is then treated as a ground-truth HMM.  The goal then is to estimate another HMM $\hat{\Theta}$ from the samples generated from the ground-truth HMM $\Theta$.
Note that the ``ground-truth HMM" is itself an estimate from a larger collection of data. However, here we are interested in how an HMM estimated from data compares with the original MM used to generate the data, when the number of samples and the sequence lengths vary.

\subsection{Estimating HMMs}

Given a ground-truth HMM $\Theta$, a set of length-$T$ fixation sequences $\{\bx\s{n}\}_{n=1}^N$ are sampled from the HMM according to the probability distribution in \refeqn{eqn:HMMjoint}.  Next an HMM $\hat{\Theta}$ is learned from the samples using variational Bayesian estimation, as implemented in EMHMM. The hyperparameters of the model, including the number of states, are estimated automatically by maximizing the marginal likelihood of the data. Hence, it is possible that the predicted number of states $\hat{K}$ is different from the number of states $K$ in the ground-truth HMM.

\begin{figure}[tbhp]
 \small
  \centering
  \includegraphics[width=1.0\textwidth]{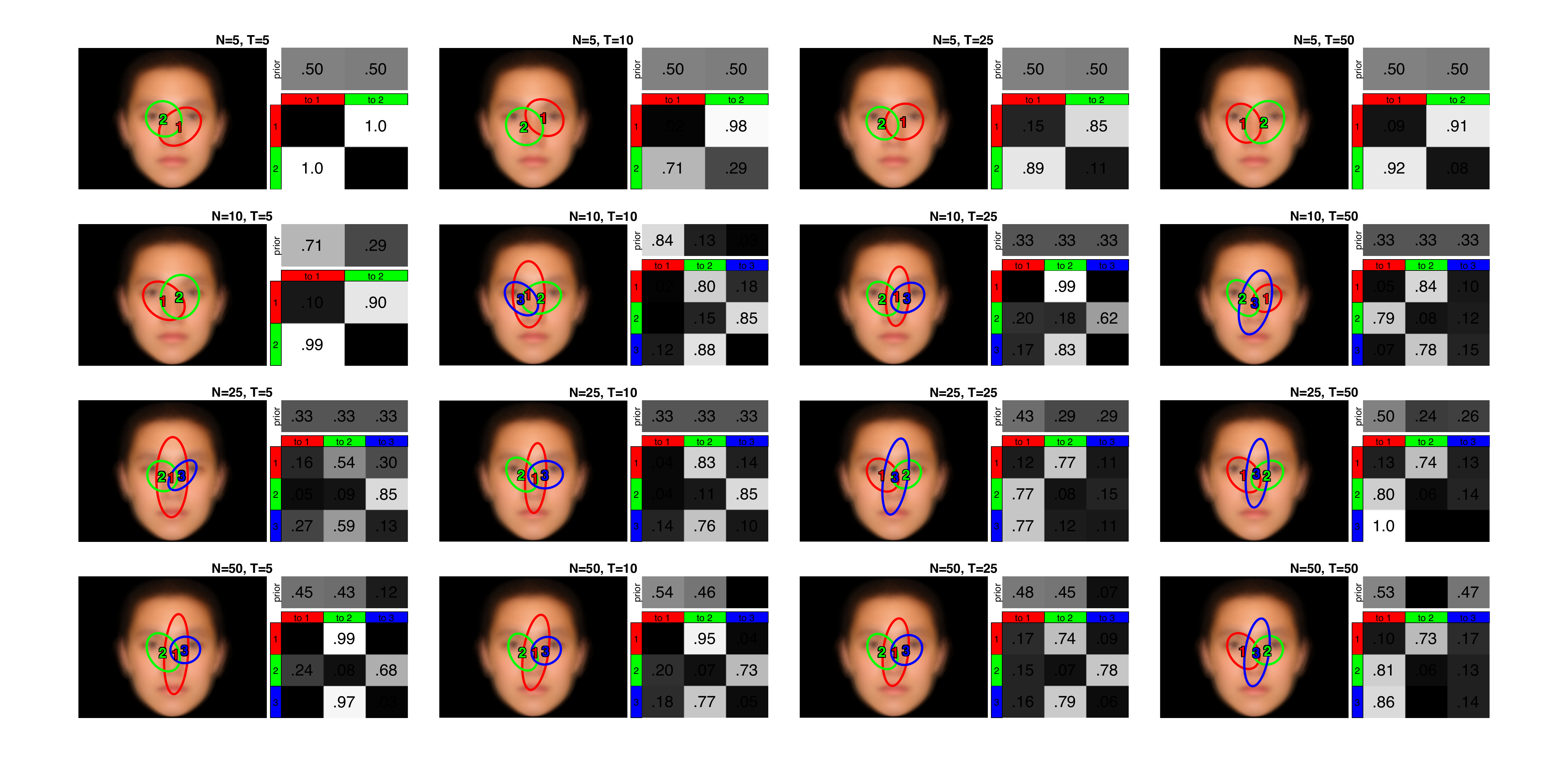} 
  \caption{Estimated HMMs for different number of sequences $N$ and sequence lengths $T$. In general, the estimation quality increases as $N$ and $T$ increase.}
  \label{fig:hmmsNT}
\end{figure}

Figure \ref{fig:hmmsNT} plots an example of HMMs estimated from sample sets for various values of $(N,T)$. In general, the estimation quality increases as $N$ and $T$ increase. 

\subsection{Comparing HMMs}

The HMM is a probabilistic model for a whole sequence $\bf{x}$, as in \refeqn{eqn:HMMobs}. 
 It also contains different distributions, including observation likelihoods, transition probabilites, and initial state probabilities in \refeqn{eqn:HMMinitial}, \refeqn{eqn:HMMtrans}, and \refeqn{eqn:HMMlik}.  Hence, we consider different methods to compare two HMMs, $\Theta$ and $\hat{\Theta}$.

\subsubsection{Dissimilarity Measures}
We first introduce the two dissimilarity measures that we will use. Consider two probability distributions $p(y)$ and $\hat{p}(y)$, we consider two methods for comparing them.

\subsubsubsection{Kullback Leibler divergence (KLD)}
Kullback Leibler divergence \citep{kullback1997information} is a dissimilarity measure between two distributions defined as 
	\begin{align}
	\phi(p(y), \hat{p}(y)) = \int_{y} p(y) \log \frac{p(y)}{\hat{p}(y)} dy = \EV_{y\sim p(y)} \left[ \log \frac{p(y)}{\hat{p}(y)}  \right],
	\label{eqn:KLD}
	\end{align}
 The KLD will be 0 when the two distributions are equal to each other, and more positive values indicate more dissimilarity between the two distributions.  Intuitively, the KLD is the amount of ``information'' that needs to be added to turn the approximation $\hat{p}(\bx)$ into the original $p(\bx)$.  Another interpretation is that the KLD is the weighted average of the log-likelihood ratio between the two models over all sequences, where the weight for each sequence is based on its likelihood of occurring.
 
In some cases, there is no analytical solution to the integral in \refeqn{eqn:KLD}, and hence we use sampling to approximate the KLD:
	\begin{align}
	\phi(p(y), \hat{p}(y))  \approx \frac{1}{S} \sum_{s=1}^{S} \log \frac{p(y\s{s})}{\hat{p}(y\s{s})},
	\label{eqn:KLDsamp}
	\end{align}
where the samples are drawn from $p(y)$, i.e., $y\s{s} \sim p(y)$.

\subsubsubsection{L1-norm} 
The L1-norm measures the absolute difference between the two distributions,
	\begin{align}
	\psi(p(y), \hat{p}(y)) = \frac{1}{2} \int_y | p(y) - \hat{p}(y) | dy.
	\end{align}
Intuitively, the L1-norm measures the amount of distribution in $\hat{p}(y)$ that needs to be moved to change $\hat{p}(y)$ into $p(y)$.  It is also is  inversely related to the histogram intersection, which measures the amount of overlapping distribution between $p(y)$ and $\hat{p}(y)$, 
	\begin{align}
	\int_y \min(p(y), \hat{p}(y)) dy = 1- \psi(p(y), \hat{p}(y)) 
	\end{align}
When the integral does not have a closed-form solution, then we approximate using numeric integration.

Because it has a more intuitive interpretation related to the percentage of overlap between two probability distributions, we will use L1-norm measurements to compare individual components of the HMM, including the individual ROIs, transition matrices, and initial state probabilities.

\subsubsection{Whole HMM}

To compare the whole HMMs, we use the Kullback-Leibler divergence rate \citep{kullback1997information} between the observation sequence likelihoods in \refeqn{eqn:HMMobs}.  Denote $p(\bx)$ and $\hat{p}(\bx)$ as the observation sequence likelihood for the ground-truth HMM $\Theta$ and the estimated HMM $\hat{\Theta}$.  The KLD rate is
	\begin{align}
	D_{HMM} = \frac{1}{T} \phi(p(\bx), \hat{p}(\bx)),
	\label{eqn:KLDobs}
	\end{align}
where $T$ is the sequence length (same as the sequence length of the data). 

\begin{figure}[tbhp]
 \small
  \centering
  \includegraphics[width=1.0\textwidth]{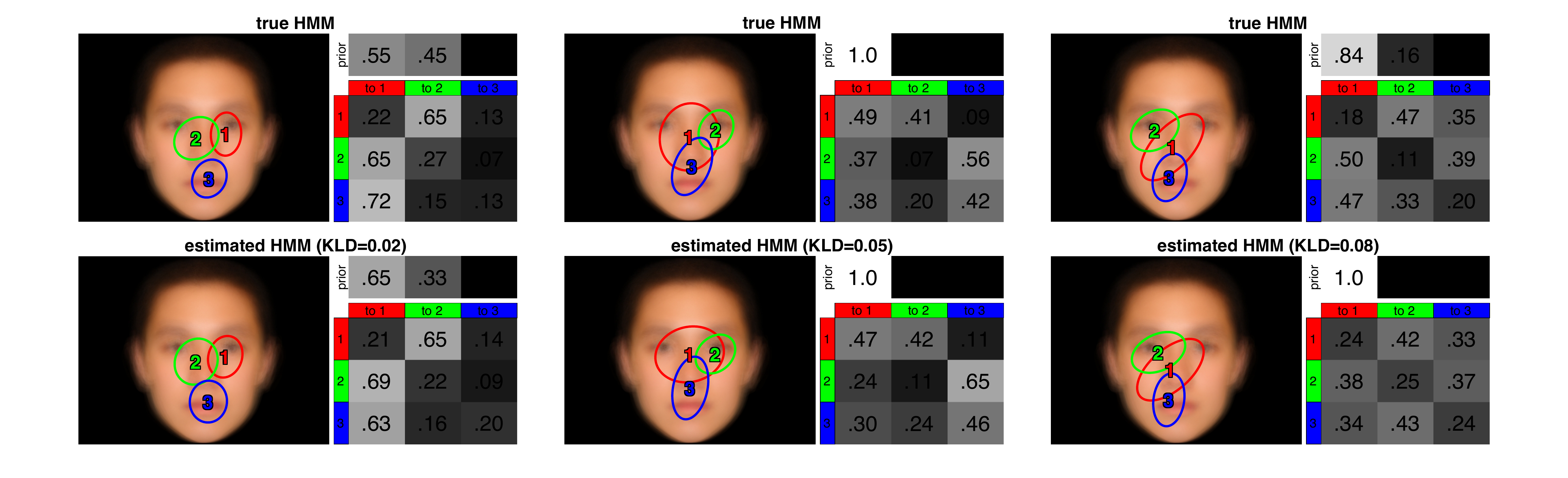} 
  \caption{True and estimated HMMs that differ by (left) KLD of 0.02, (center) KLD of 0.05, and (right) KLD of 0.08.}
  \label{fig:eg_hmms_KL}
\end{figure}

Figure \ref{fig:eg_hmms_KL} plots an examples of the KL divergence between a few true and estimated HMMs.  For KLD values of 0.05 and less, the estimated HMMs are close to the true HMMs, i.e., the reflect the same strategy.

\CUT{
\subsubsection{Observation distributions}

Besides comparing distributions of the whole sequence of observations, we can also compare distributions of observations at a single time. Here we compare the initial observation distribution $p(\bx_1)$, which represents to the first fixation, and the conditional distribution $p(\bx_t|\bx_{t-1})$, which represents the transition between fixation locations.

\subsubsubsection{Initial Observation}
We use the L1 norm to compare the initial observation distributions,
	\begin{align}
	L_{prior} = \psi(p(\bx_1), \hat{p}(\bx_1))
	\end{align}

\subsubsubsection{Conditional Observation}
We use the conditional KLD to compare the conditional observation distributions at steady-state, 
	\begin{align}
	D_{cond} = \EV_{\bx_{t-1}} \left[ \phi(p^*(\bx_t|\bx_{t-1}), \hat{p}^*(\bx_t|\bx_{t-1})) \right], 
	\end{align}
which is the averaged KLD of the conditional observation distribution, averaged over the probable previous fixations $\bx_{t-1}$.

Using a few properties of KLD, we can derive this as
	\begin{align}
	D_{cond} =  \phi(p^*(\bx_t, \bx_{t-1}), \hat{p}^*(\bx_t, \bx_{t-1})) - \phi(p^*( \bx_{t-1}), \hat{p}^*(\bx_{t-1})).
	\end{align}
}

\subsubsection{Matched ROIs}

The above comparisons measure differences in the distributions of the whole sequences between two HMMs, considering all ROIs and transitions at the same time.  We can also compare the individual ROIs between the two HMMS by matching ROIs between $\Theta$ and $\hat{\Theta}$.  For now assume that both HMMs have the same number of ROIs.  Assuming that the ROI indices already match, the L1-norm between the two HMMs $\Theta$ and $\hat{\Theta}$ is
	\begin{align}
	\Psi(\Theta, \hat{\Theta}) = \frac{1}{K} \sum_{j=1}^K \psi( p(\bx_t|z_t=j), \hat{p}(\bx_t|z_t=j))
	\end{align}
When the ROIs do not match, then we define $\calP$ as the state-permutation operator that permutes (reorders) the states of an HMM, and compute  the permutation that minimizes the L1-norm
	\begin{align}
	\ell_{ROI}= \min_{\calP} \Psi(\Theta, \calP(\hat{\Theta})).
	\end{align}
When the number of ROIs do not match between $\hat{\Theta}$ and $\Theta$, e.g., $\hat{K} < K$, then we duplicate some ROIs in $\hat{\Theta}$ until $\hat{K}=K$.  In this way, the matching function will match one ROI in $\hat{\Theta}$ to more than one ROI in $\Theta$.  The ROIs for duplication are selected so as to minimize the final $\ell_{ROI}$.  A similar procedure occurs if $\hat{K} > K$.

\begin{figure}[tbhp]
 \small
  \centering
  \includegraphics[width=1.0\textwidth]{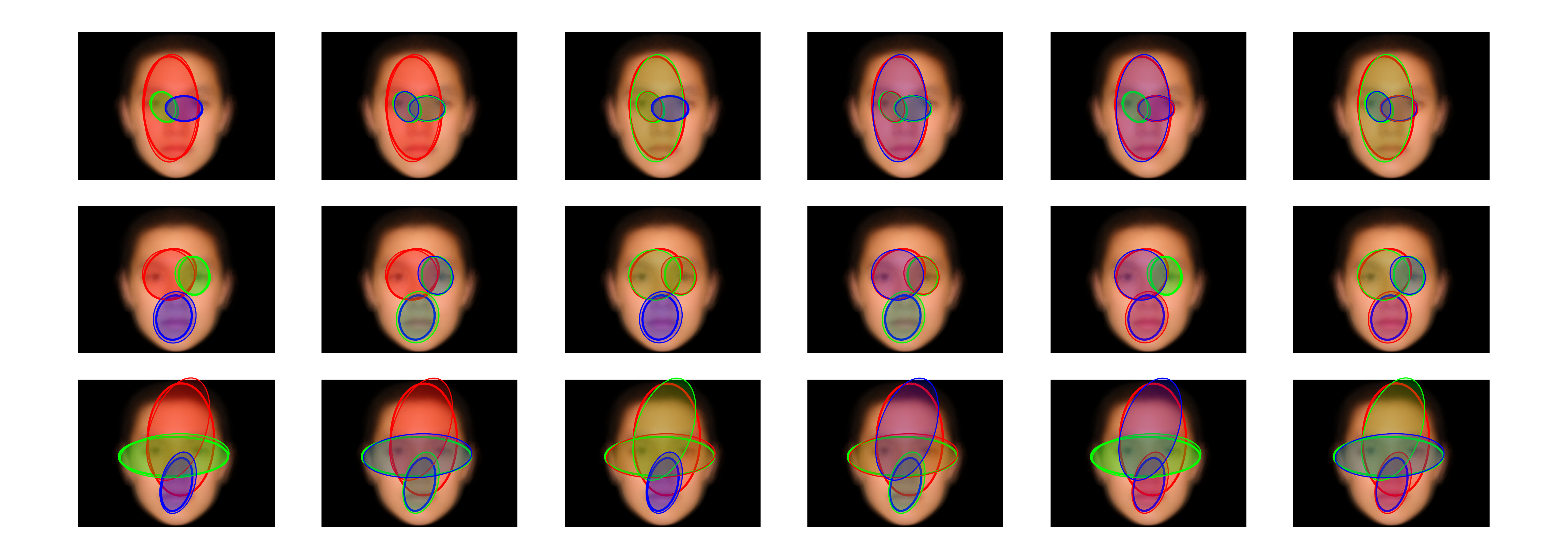} 
  \caption{Permutations of labels of estimated ROIs to best match the true ROIs. Each row shows one set of estimated/true ROIs.  The thin line is the estimated ROI, and the thick line is the true ROI. Colors indicate ROI labels.  The best match is the left-column.}
  \label{fig:eg_perm_roi}
\end{figure}

Figure \ref{fig:eg_perm_roi} plots examples of permuting the labels of the ROIs for the estimated HMM to best match the true HMM ROIs.  

\begin{figure}[tbhp]
 \small
  \centering
  \includegraphics[width=1.0\textwidth]{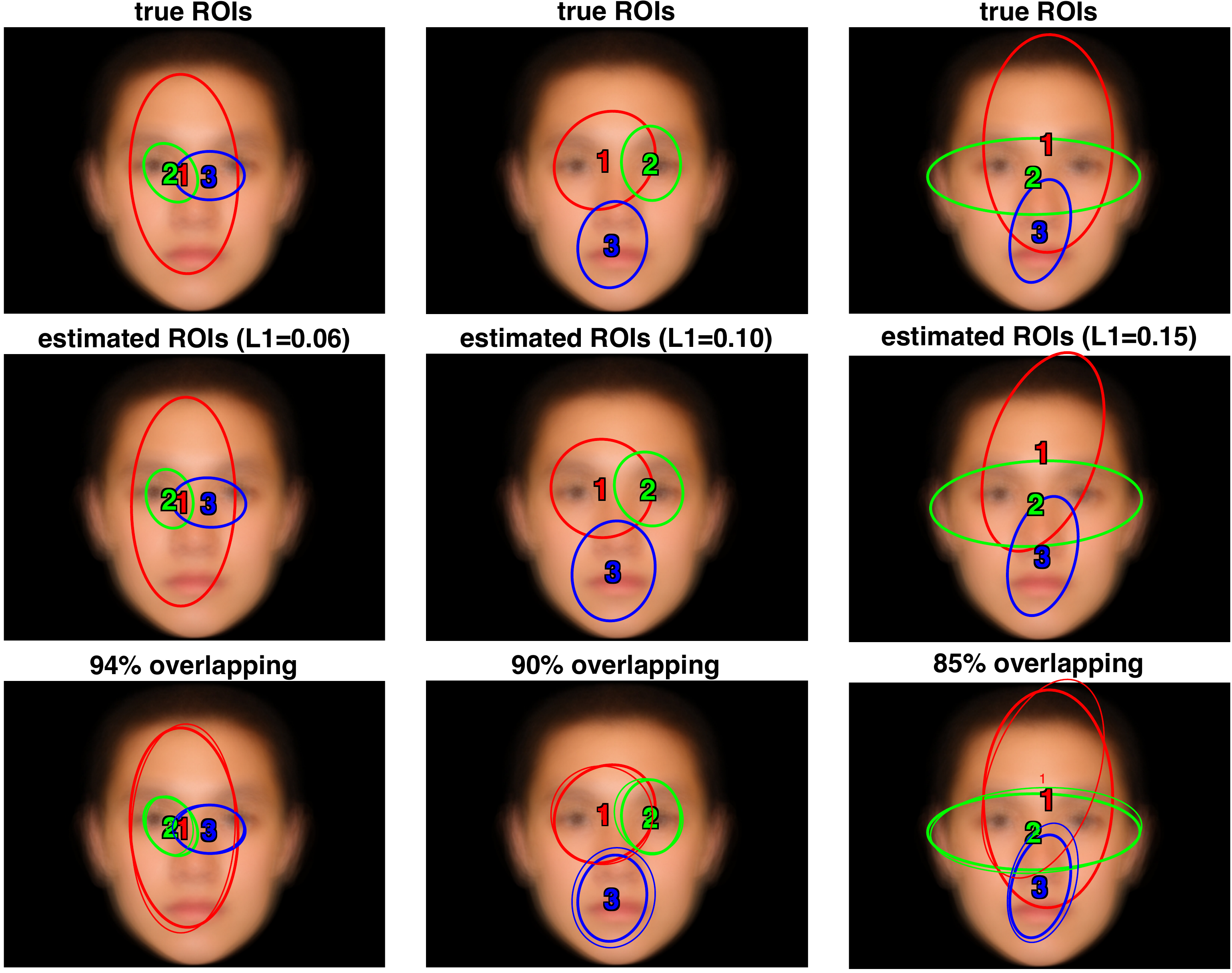} 
  \caption{True and estimated ROIs that differ by (left) L1 of 0.06, (center) L1 of 0.10, and (right) L1 of 0.15.}
  \label{fig:eg_ROI_L1}
\end{figure}

Figure \ref{fig:eg_ROI_L1} plots an examples of the matched L1-norm between the true and estimated ROIs.   The contours of the true and estimated ROIs are well matched for L1-norm of 0.10, which corresponds to 90\% overlapping. 

\subsubsection{Matched Transition Matrices and Initial States}

We directly compare transition matrices and initial state probabilites using a similar matching technique.  Here we are interested in whether the underlying state dynamics (transitions and prior) are similar, regardless of the accuracy of the ROIs.  For now assume that the number of states it the same in the two HMMs $\Theta$ and $\hat{\Theta}$.
	Define the L1-norm between two transition matrices, 
	\begin{align}
	\Psi(\bA, \hat{\bA}) = \sum_{i=1}^K \sum_{j=1}^K | a_{i,j} - \hat{a}_{i,j}|, 
	\end{align}
	and the L1-norm between two initial states probability vectors,
	\begin{align}
	\Psi(\bpi, \hat{\bpi}) = \sum_{j=1}^K |\pi_{j} - \hat{\pi}_{j}|.
	\end{align}
We then find the permutation that minimizes the L1-norm between the transition matrices and priors,
	\begin{align}
	\calP^* = \min_{\calP} \Psi(\bA, \calP(\hat{\bA})) + \Psi(\bpi, \calP(\hat{\bpi})).
	\label{eqn:minperm}
	\end{align}
Given the optimal permutation $\calP^*$, we compute L1-norms for comparison:
	\begin{align}
	\ell_{trans} &= \frac{1}{K}\Psi(\bA, \calP^*(\hat{\bA})),
	\\
	\ell_{prior} &= \Psi(\bpi, \calP^*(\hat{\bpi})).
	\end{align}
	
When the number of ROIs do not match between $\hat{\Theta}$ and $\Theta$, i.e., $\hat{K}<K$, we augment $\hat{\Theta}$ with enough states to match $\Theta$.  The new state is made functionally identical to one of the original states.  This is performed by splitting the transition probabilities to the old state between the new and the old state, and setting the transition matrix row of the new state to be the same as that of the old state.  In particular, let $j'$ be the new state and $j$ be the old state, then we set the new transition probabilites $\tilde{a}$ as
	\begin{align}
	\tilde{a}_{i,j} &= \tilde{a}_{i,j'} = \tfrac{1}{2}a_{i,j}, \quad \forall i\neq j'
	\\
	\tilde{a}_{j',k} &= \tilde{a}_{j,k}, \quad \forall k.
	\end{align}
Similarly, the initial probability for state $j$ is split with state $j'$, to form the new initial probabilities,
	\begin{align}
	\tilde{\pi}_{j} &= \tilde{\pi}_{j'} = \tfrac{1}{2}\pi_{j},
	\end{align}
The old state $j$ for duplication is selected to minimize the permutation error in \refeqn{eqn:minperm}.  A similar procedure occurs if $\hat{K} > K$.

\begin{figure}[tbhp]
 \small
  \centering
  \includegraphics[width=1.0\textwidth]{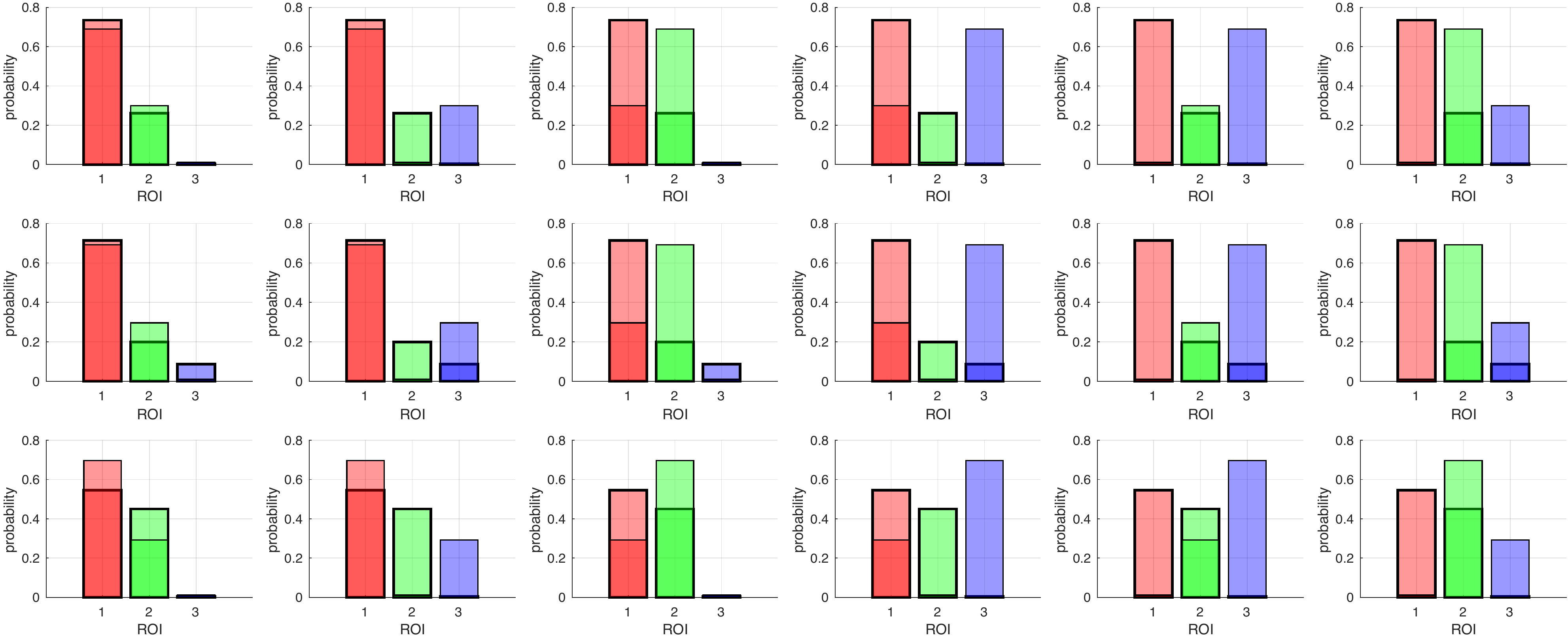} 
  \caption{Permutations of prior probabilities to best match the true ones. Each row shows one set of estimated/true priors.  The thin line is the estimated prior, and the thick line is the true prior. Colors indicate state labels.  The best match is the left-column.}
  \label{fig:eg_perm_prior}
\end{figure}

Figure \ref{fig:eg_perm_prior} plots examples of permuting the prior probabilities for the estimated HMM to best match those of the true HMM.

\begin{figure}[tbhp]
 \small
  \centering
  \includegraphics[width=1.0\textwidth]{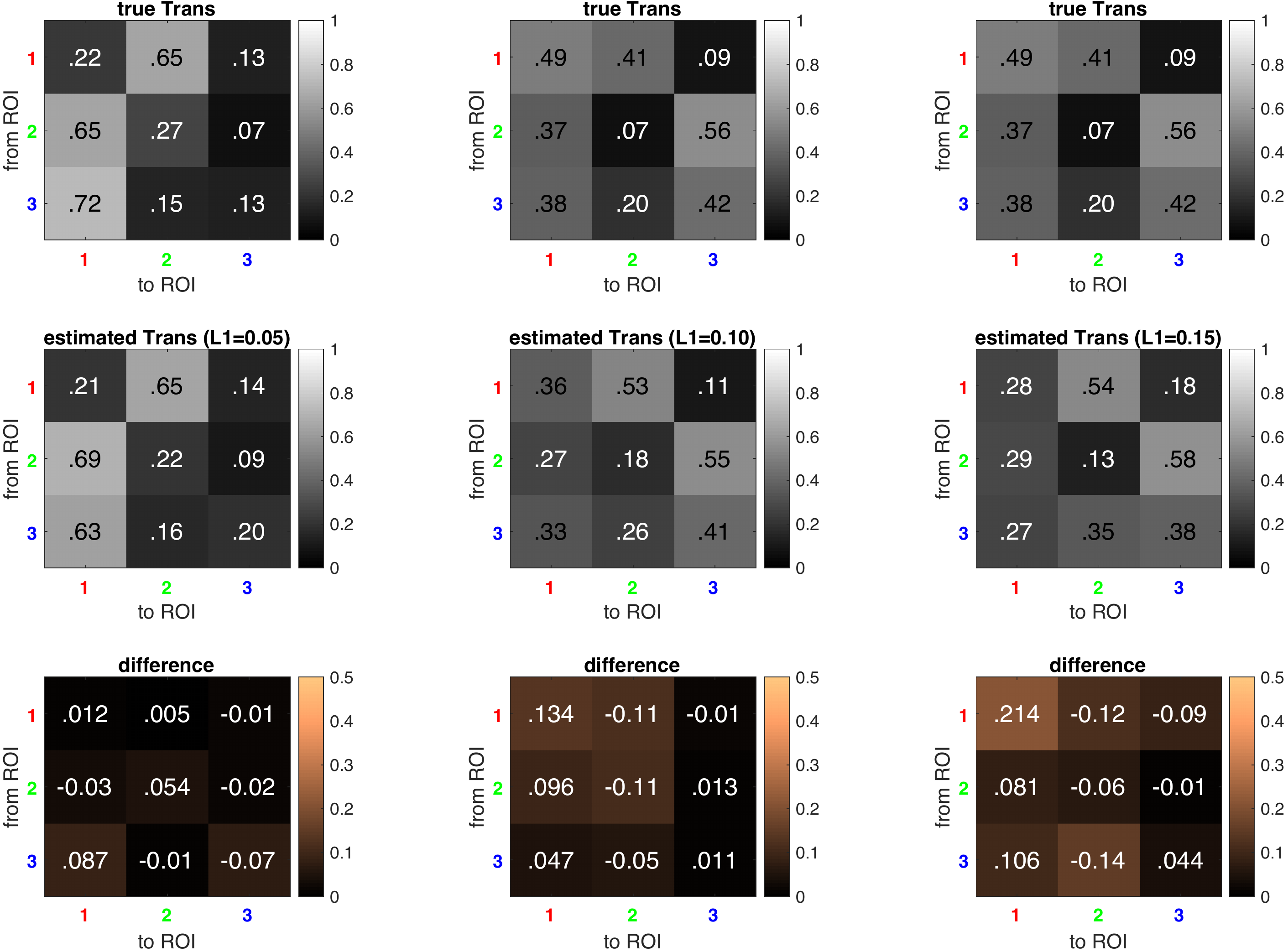} 
  \caption{True and estimated transition matrices that differ by (left) L1 of 0.05, (center) L1 of 0.10, and (right) L1 of 0.15.}
  \label{fig:eg_trans_L1}
\end{figure}

\begin{figure}[tbhp]
 \small
  \centering
  \includegraphics[width=0.8\textwidth]{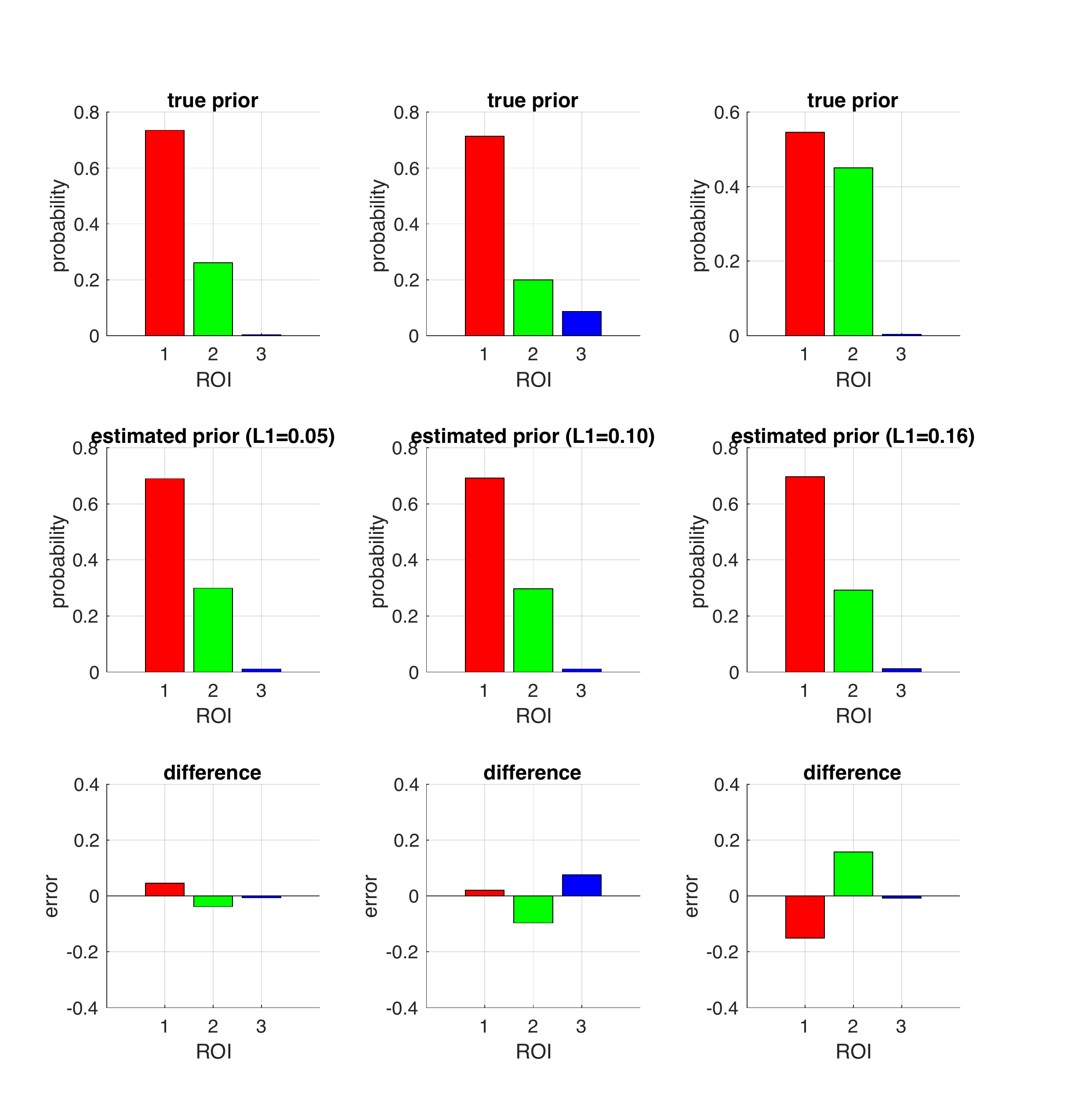} 
  \caption{True and estimated prior probabilities that differ by (left) L1 of 0.05, (center) L1 of 0.10, and (right) L1 of 0.15.}
  \label{fig:eg_prior_L1}
\end{figure}

 Figures \ref{fig:eg_trans_L1} and \ref{fig:eg_prior_L1} plot examples of L1-norm between discrete probability distributions, i.e., the transition matrix and the prior probabilities. In this case, L1 of 0.10 corresponds to needing to move 10\% probability mass to make the estimated distribution into the true distribution.


\section{Experiment Setup}
\label{text:setup}

In the experiments, we use 10 subjects to create 10 ground-truth HMMs.  The stimuli images are 512$\times$384, with the face region roughly 300$\times$350.
We consider different numbers of samples $N$ and sequence lengths $T$.
For each combination of $(N,T)$, we run 500 trials, where each trial randomly selects one ground-truth HMM $\Theta$, samples $N$ sequences of length $T$ from which an HMM $\hat{\Theta}$ is estimated.  The ground-truth and estimated HMMs are then compared using the metrics described above:
sequence KLD ($D_{HMM}$), 
\CUT{prior L1-norm ($L_{prior}$), conditional KLD ($D_{cond}$), }
matched L1-norm for ROIs ($\ell_{ROI}$), matched L1-norm for transitions ($\ell_{trans}$), and matched L1-norm for priors ($\ell_{prior}$).

For comparison, we also add known distortions to the parameters of a ground-truth HMM to obtain  noisy HMM $\tilde{\Theta}$, and then compute the same metrics.  This allows calibration of the metrics to known distortion of the HMM parameters.
In particular, we consider four types of distortion (deviation):
\begin{itemize}
\item {\bf ROI Mean:} move the ROI mean by a fixed distance $\alpha$: $\tilde{\bmu}_j = \mu_j + \alpha \br$, where $\br$ is a random unit-length vector.
\item {\bf ROI Covariance:} increase or decrease the size of the covariance matrix by $\beta$. Let $\Sigma=V\Lambda V^T$ be the eigen-decomposition. Then the covariance matrix is scaled as $\tilde{\Sigma} = V \Lambda^{(1+\beta r)} V^T$, where random value $r\in\{-1,1\}$ and $\beta$ is the deviation parameter.
\item {\bf Prior Vector:} shift probability mass between states such that the L1-norm is $\delta$, i.e., $\psi(\bpi, \tilde{\bpi})=\delta$. Specifically, $\tilde{\bpi} = \bpi + \frac{2}{\delta \bone^T|\br|} \br$ where $\br$ is a random vector that makes $\tilde{\bpi}$ into a valid probability distribution.
\item {\bf Transition Matrix:}  for each row of the transition matrix, shift probability mass between states such that the L1-norm is $\epsilon$, i.e., $\psi(\ba_j, \tilde{\ba_j})=\epsilon$. Specifically, $\tilde{\ba}_j = \ba_j + \frac{2}{\epsilon \bone^T|\br|} \br$, where $\br$ is a random vector that makes $\tilde{\ba}_j$ into a valid probability distribution.
\end{itemize}
We compute the error metrics for HMMs using different values of the distortion parameters $\{\alpha, \beta, \delta, \epsilon\}$, averaged over 500 trials.  Then, we can interpret a particular level of error as equivalent to the corresponding level of distortion in one of the HMM parameters.

\section{Experiment Results}
\label{text:results}

We next present the experiment results of the simulation study.

\subsection{Comparison of whole HMMs}

\begin{figure}[tbhp]
 \small
  \centering
  \begin{tabular}{cc}
 \raisebox{5cm}{(a)} \includegraphics[width=0.45\textwidth]{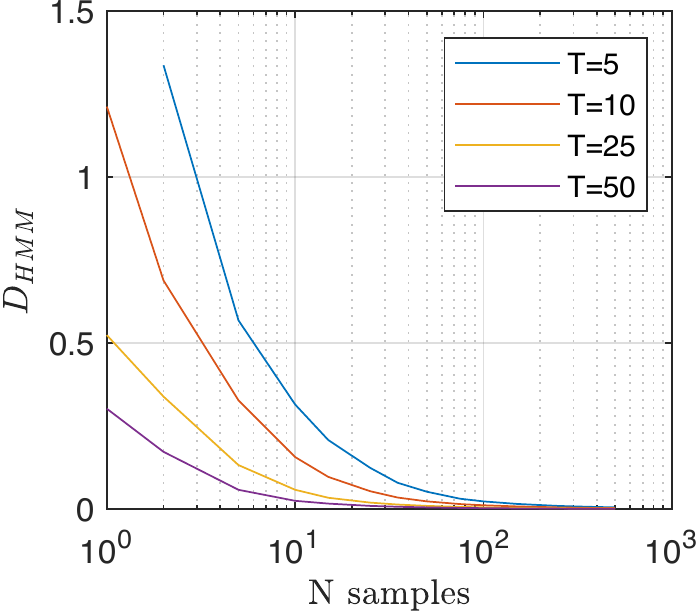} &
  \\
  \raisebox{3cm}{(b)} \includegraphics[width=0.45\textwidth]{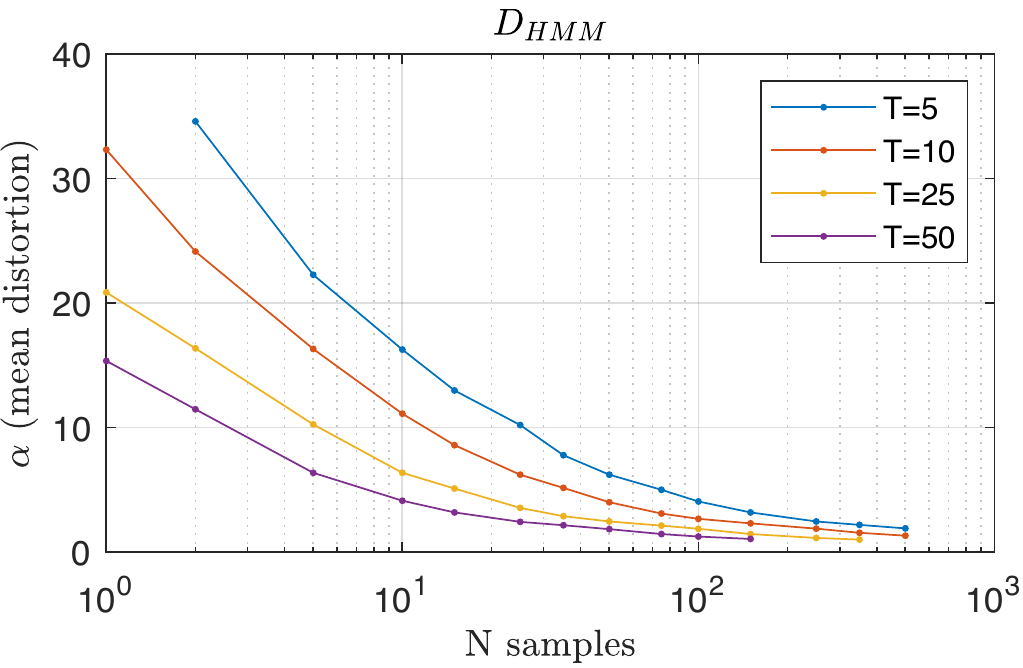} &
  \raisebox{3cm}{(c)} \includegraphics[width=0.45\textwidth]{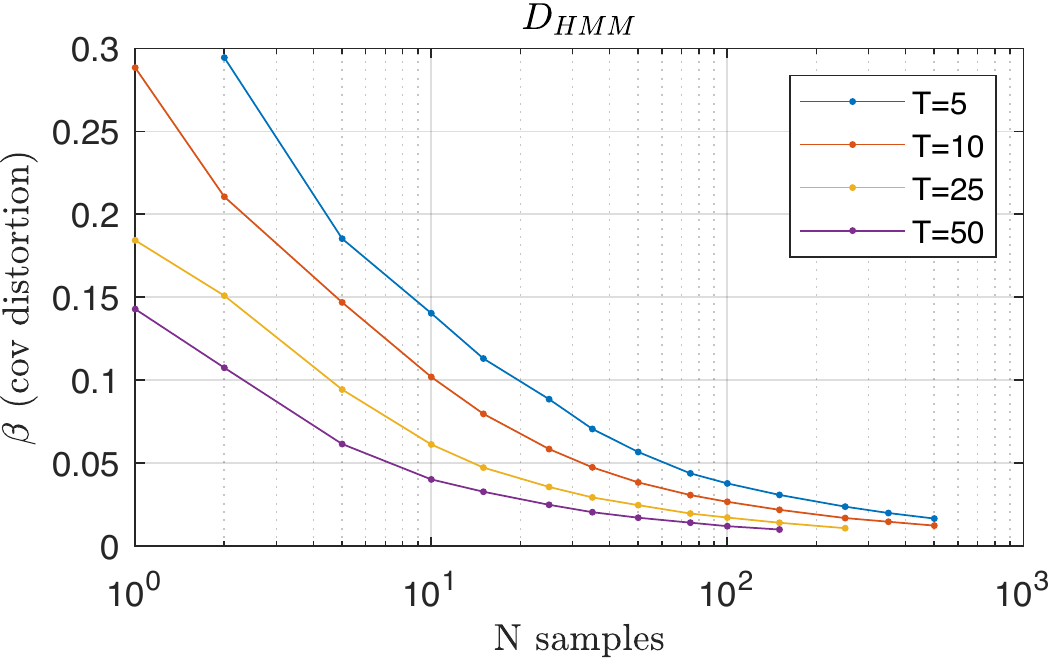}
  \\  
  \raisebox{3cm}{(d)} \includegraphics[width=0.45\textwidth]{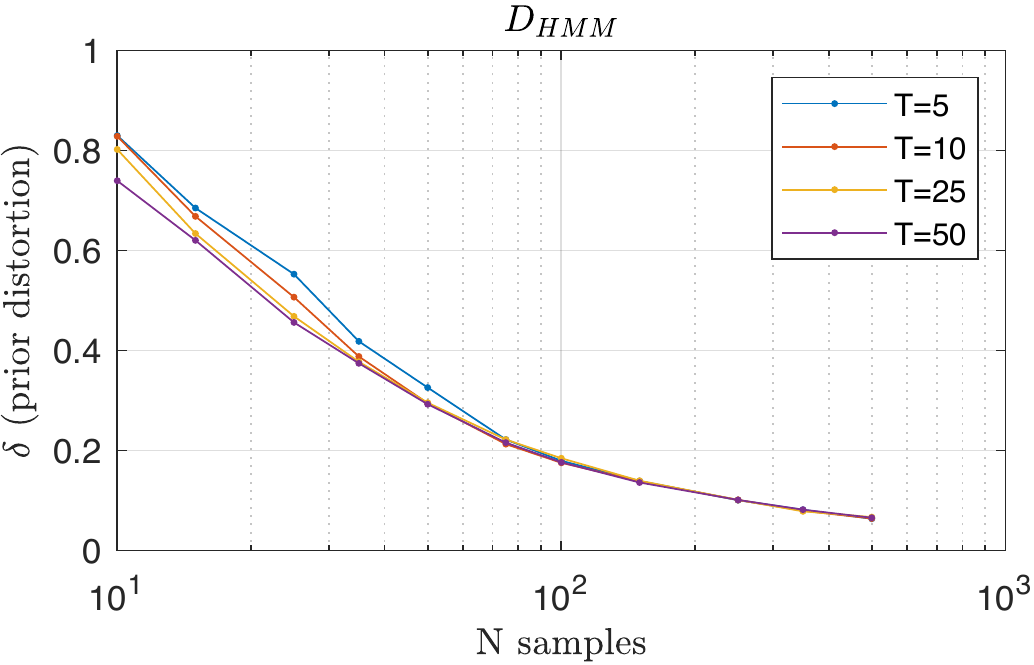} &
  \raisebox{3cm}{(e)} \includegraphics[width=0.45\textwidth]{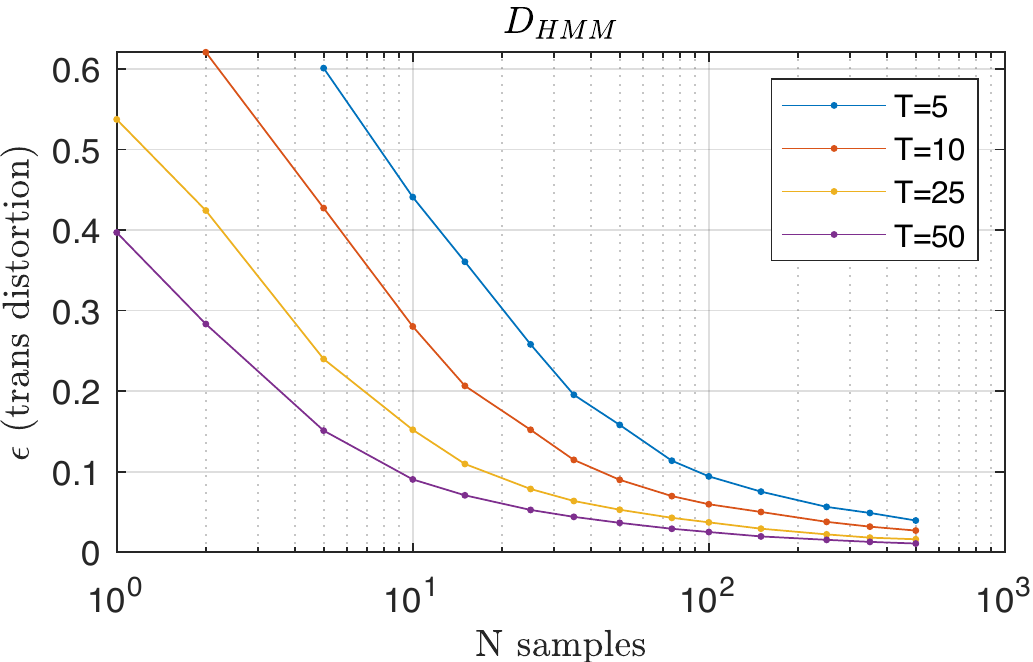}
  \end{tabular}
  \caption{(a) whole sequence KL divergence ($D_{HMM}$) versus number of sequences $N$ and sequence length $T$, and its equivalence to known distortion in the HMM parameters (b) mean, (c) covariance, (d) prior, (e) transition matrix.}
  \label{fig:D_HMM}
\end{figure}

The results for the KLD for whole HMMs is shown in Figure \ref{fig:D_HMM}.
The KLD decreases as the number of samples increases or the length of the sequences increases, and eventually converges to zero.  To obtain a low KLD of 0.05 requires roughly 250 individual fixations, e.g., 52 length-5 sequences, 26 length-10 sequences, 11 length-25 sequences or 6 length-50 sequences. 

%Looking at the equivalent parameter distortion,
%in order to obtain the ROI mean within 10 pixels requires 30 length-5 samples, 10 length-10 samples, 5 length-25 samples, or 3 length-50 samples.
%In order to obtain the covariance matrix within 5\% requires 60 length-5 samples, 35 length-10 samples, 15 length-25 samples, or 7 length-50 samples.

\CUT{
\subsection{Comparisons of observation distributions}

\begin{figure}[tbhp]
 \small
  \centering
  \begin{tabular}{cc}
 \raisebox{5cm}{(a)} \includegraphics[width=0.45\textwidth]{figures/D_cond.pdf} &
 \raisebox{5cm}{(c)} \includegraphics[width=0.45\textwidth]{figures/L_prior.pdf} 
  \\
  \raisebox{3cm}{(b)} \includegraphics[width=0.45\textwidth]{figures/D_cond_trans.pdf} &
  \raisebox{3cm}{(d)} \includegraphics[width=0.45\textwidth]{figures/L_prior_combo.pdf}
  \end{tabular}
  \caption{(a) conditional KL divergence ($D_{cond}$) versus number of sequences $N$ and sequence length $T$, and (b) its equivalence to known distortion of the transition matrix; 
  (c) L1-norm of prior ($L_{prior}$), and (d) its equivalence to known distortion of the prior vector.}
  \label{fig:obs}
\end{figure}
}

\subsection{Comparison of matched ROIs}

\begin{figure}[tbhp]
 \small
  \centering
  \begin{tabular}{cc}
 \raisebox{5cm}{(a)} \includegraphics[width=0.45\textwidth]{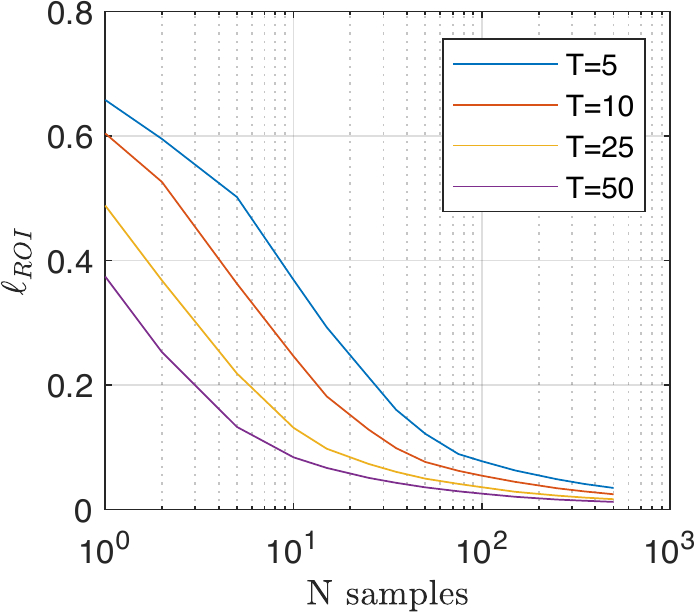}  &
  \\
  \raisebox{3cm}{(b)} \includegraphics[width=0.45\textwidth]{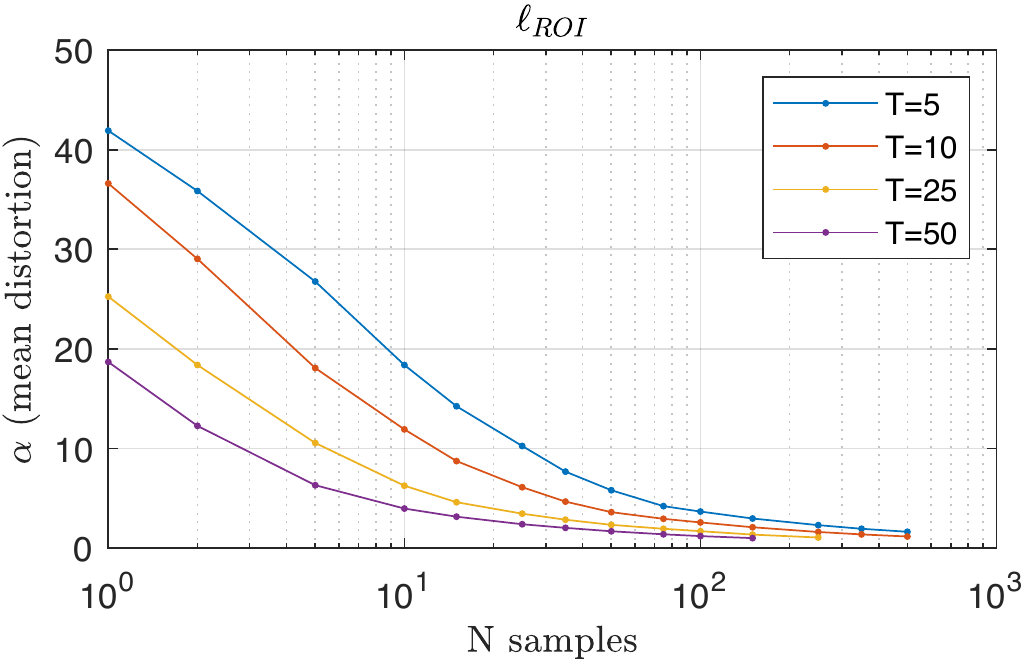} &
  \raisebox{3cm}{(c)} \includegraphics[width=0.45\textwidth]{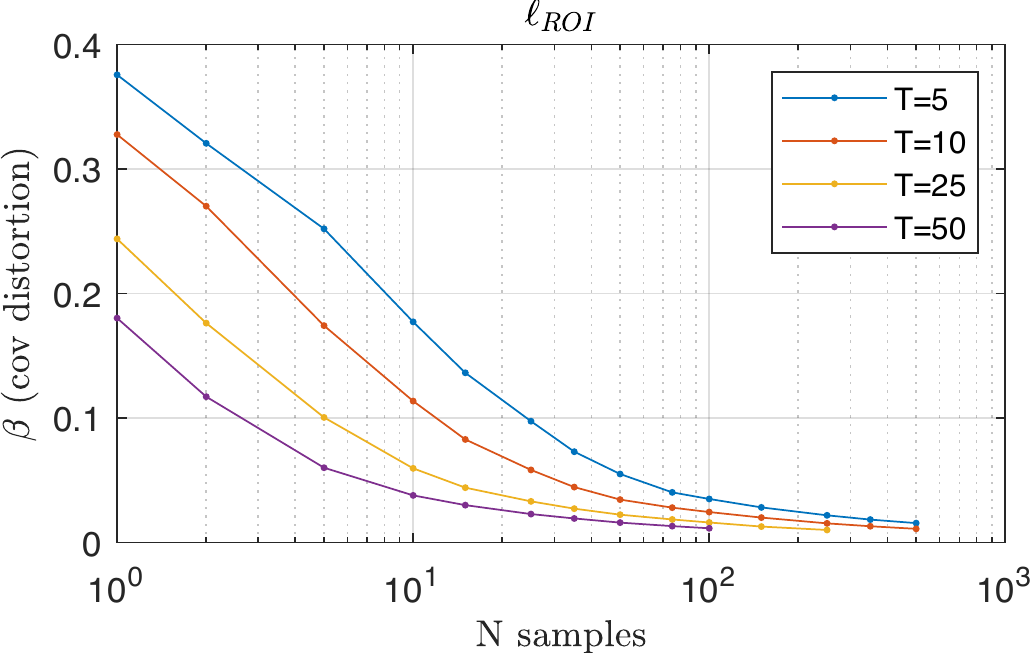}
  \end{tabular}
  \caption{(a) L1-norm between matched ROIs ($\ell_{ROI}$), and its equivalence to known distortion in the (b) ROI mean and (c) covariance matrix. }
  \label{fig:ROI}
\end{figure}

Figure \ref{fig:ROI} shows the L1-norm between matched ROIs and its equivalent distortions in the mean and covariance parameters of the ground-truth ROI.
To obtain 90\% overlap (10\% L1-norm) of ROIs between the two HMMs requires roughly 350 individual fixations (e.g., 66 length-5 sequences, 35 length-10 sequences, 15 length-25 sequences or 8 length-50 sequences).  The total number of fixations is important since the ROIs are determined using all of the fixations from the samples.  Here,  90\% overlap from 350 individual fixations (e.g., 35 length-10 sequences) corresponds to roughly 4.7 pixel error in the mean or 4.5\% change in the size of the ROI, as seen in Figures \ref{fig:ROI}(b) and \ref{fig:ROI}(c).

\subsection{Comparison of matched states}

\begin{figure}[tbhp]
 \small
  \centering
  \begin{tabular}{cc}
 \raisebox{5cm}{(a)} \includegraphics[width=0.45\textwidth]{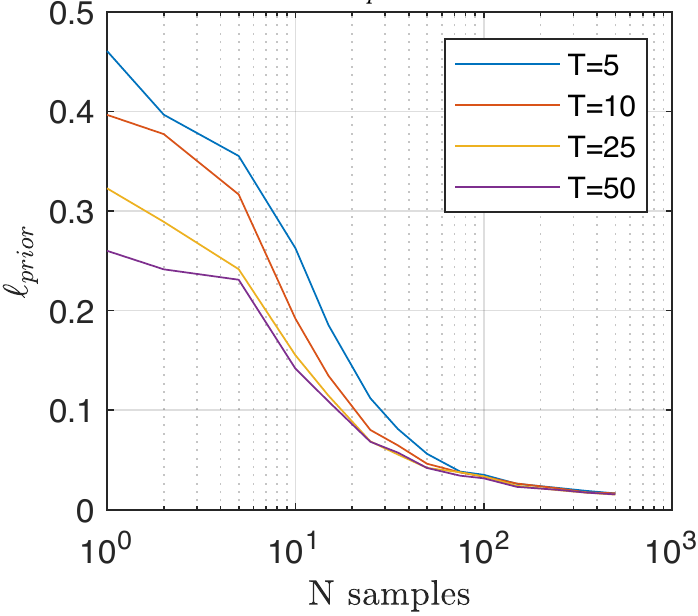} &
 \raisebox{5cm}{(c)} \includegraphics[width=0.45\textwidth]{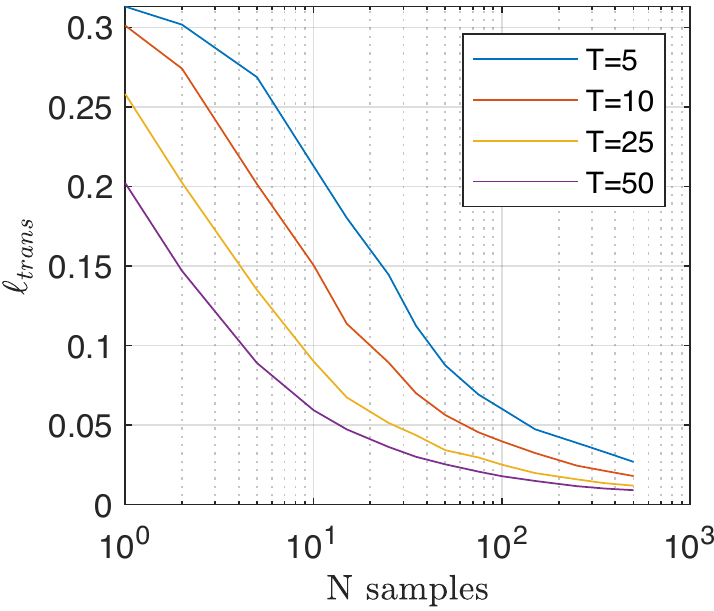} 
  \\
  \raisebox{3cm}{(b)} \includegraphics[width=0.45\textwidth]{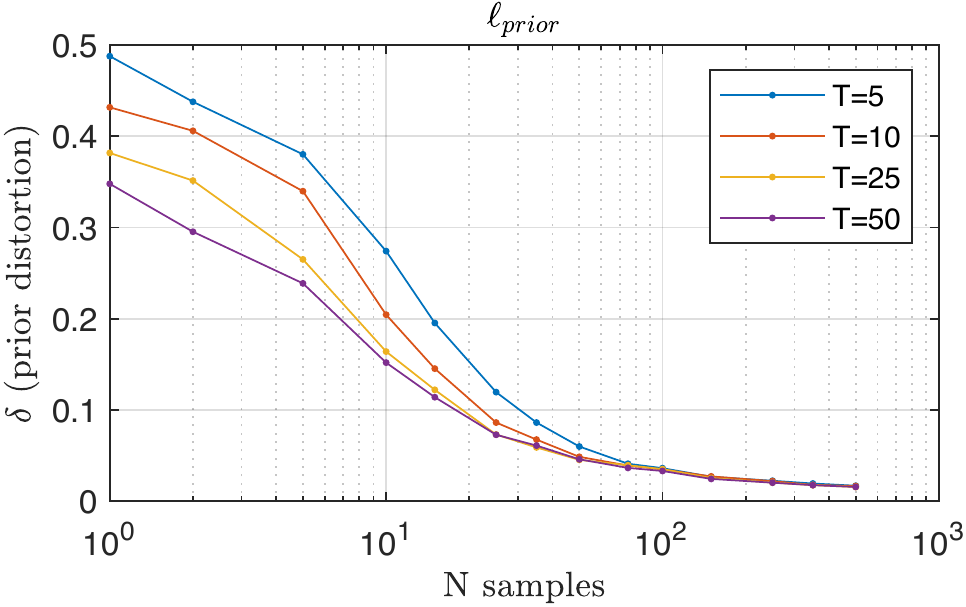} &
  \raisebox{3cm}{(d)} \includegraphics[width=0.45\textwidth]{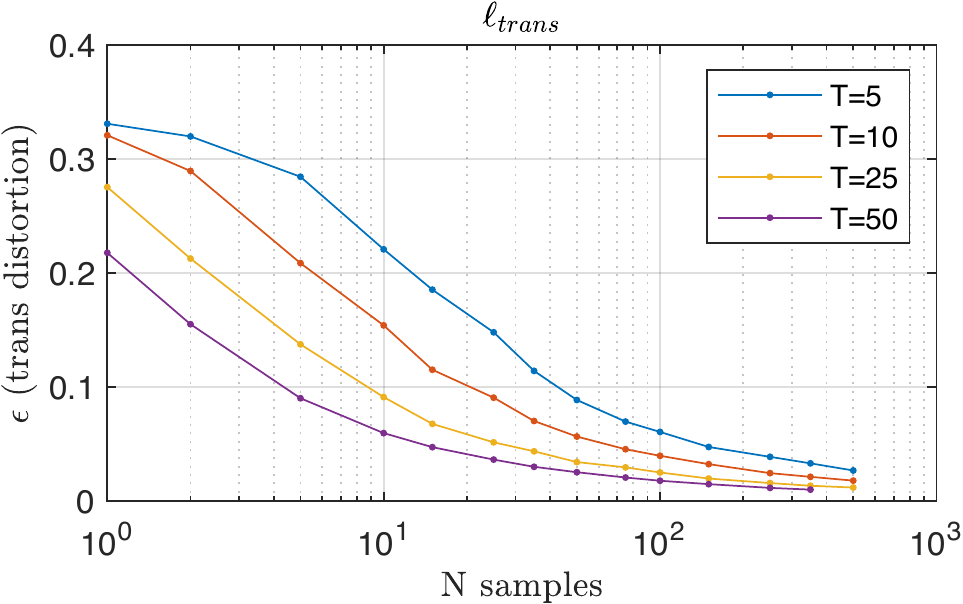}
  \end{tabular}
  \caption{(a) L1-norm of the matched prior states ($\ell_{prior}$)  versus number of sequences $N$ and sequence length $T$, and (b) its equivalence to known distortion of the prior vector; 
  (c) L1-norm of matched transition matrices ($\ell_{trans}$), and (d) its equivalence to known distortion of the transition matrix.}
  \label{fig:states}
\end{figure}

Figure \ref{fig:states} shows the L1-norm for matched states for the prior and the transition matrix, along with the equivalent distortions to the ground-truth prior and transition matrices.  To obtain 90\% overlap of the prior probability distribution requires 18-31 sequences (each sequence only has 1 first fixation). To obtain 90\% overlap of the transition probabilities requires roughly 215 individual fixations (e.g., 43 length-5 sequences, 21 length-10 sequences, 9 length-25 sequences, or 5 length-50 sequences).  Here, the total number of fixations is important since the transition matrix is computed from pairs of fixations.

\section{Summary}
\label{text:summary}

The results of the simulation study suggest that to obtain a low KL divergence of 0.05 between the estimated and the ground-truth HMMs requires at least 250 individual fixations.
Looking at the individual components of the HMM, 
the results suggest that to obtain 90\% overlap between the estimated and ground-truth Gaussians ROIs requires at least 350 individual fixations.  To obtain 90\% overlap between the transition probabilities and between the priors requires at least 215 individual fixations and 25 sequences (first fixations).
 Hence, any combination of sequence length and number of samples that can obtain this requirement should be able to obtain good estimates of the HMM that are close to the ground-truth, and thus representative of the subject's overall eye gaze strategy.

Finally, we should note that this simulation study is testing whether estimating an HMM from samples can recover the ground-truth HMM that generated the samples.  This is helpful if we want to infer the subject's overall eye fixation strategy from the HMM, i.e., the subject's underlying process that generated the observed eye fixations. 
However, from a purely analysis point-of-view, it is still valid to learn HMMs from a limited set of samples, and then use the HMMs as a representation of the observed data, e.g., in a classification task as in \cite{SMAC}.  In this case, the HMM is serving as a summarized representation of what the subject's eye gaze pattern for the given stimuli, and may not be representative of the underlying eye gaze strategy.

%%%%%%%%%%%%%%%%
\vskip 0.2in
\bibliography{refs}

\end{document}